\documentclass[12pt]{article}
\usepackage{natbib}
\usepackage{nameref}
\usepackage{cleveref}

\usepackage{amsmath}
\usepackage{amssymb}
\usepackage{graphicx}
\usepackage{float}

\usepackage{times}

\renewcommand\refname{References and Notes}

\newcounter{lastnote}

\usepackage[ruled,vlined]{algorithm2e}

\raggedright
\usepackage[aboveskip=1pt,labelfont=bf,labelsep=period,justification=raggedright,singlelinecheck=off]{caption}

\makeatletter
\renewcommand{\@biblabel}[1]{\quad#1.}
\makeatother

\date{}

\usepackage{lastpage,fancyhdr,graphicx}
\usepackage{epstopdf}
\SetKwComment{Comment}{// }{}

\begin{document}
\vspace*{0.35in}

\begin{flushleft}
{\Large
\textbf\newline{Feynman Machine: The Universal Dynamical Systems Computer}
}
\newline
\\
Eric Laukien\textsuperscript{1},
Richard Crowder\textsuperscript{2},
Fergal Byrne\textsuperscript{3,*}
\\
\bigskip
\bf Ogma Intelligent Systems Corp. \textsuperscript{1}Jupiter, FL, USA, \bf\textsuperscript{2}United Kingdom, \bf\textsuperscript{3}Dublin, Ireland
\\
\bigskip


%
%





*fergal@ogmacorp.com

\end{flushleft}
\section*{Abstract}
Efforts at understanding the computational processes in the brain have met with limited success, despite their importance and potential uses in building intelligent machines. We propose a simple new model which draws on recent findings in Neuroscience and the Applied Mathematics of interacting Dynamical Systems. The Feynman Machine is a Universal Computer for Dynamical Systems, analogous to the Turing Machine for symbolic computing, but with several important differences. We demonstrate that networks and hierarchies of simple interacting Dynamical Systems, each adaptively learning to forecast its evolution, are capable of automatically building sensorimotor models of the external and internal world. We identify such networks in mammalian neocortex, and show how existing theories of cortical computation combine with our model to explain the power and flexibility of mammalian intelligence. These findings lead directly to new architectures for machine intelligence. A suite of software implementations has been built based on these principles, and applied to a number of spatiotemporal learning tasks.  



\section*{Introduction}
The problem of understanding how the brain operates seems to become more challenging as we learn more and more of its details. One major difficulty is that the computation appears to be emergent across large populations of neurons in a region, but at the same time the details of individual synaptic connections, the ``wiring diagram'', the importance of individual and bursts of spikes, and the numerous genetically defined neuron types all seem crucial to understanding. 

On the other hand, the recent success of Deep Learning, which uses dramatically simplified models of neurons and network architectures, has given rise to the impression that the details of cortical computation are no more critical to intelligence than feathers and flapping wings are to flight.

While successful in a number of domains, it is clear even to their leading proponents that current Deep Learning concepts have fundamental limitations, and that many challenges remain in the quest for machine intelligence \citep{lecun2015deep,Goodfellow-et-al-2016-Book}. At the same time, as more is learned about how actual neurons work, a huge gap is opening between systems based on point neural models conceived in the 1940's \citep{mcculloch1943logical}, and the extraordinary complexity of even a single neuron.

A confounding problem is that the traditional mathematical tools used in both neuroscience and neural networks are ill-suited to either domain. Methods based on statistical physics were designed to model and reason about large populations of essentially opaque, elastic atomic particles, whose interactions could be described as reversible exchanges of momentum and energy. Similarly, attempts to apply ideas from information theory (which itself is a kind of abstraction of thermodynamics) have been successful only when models have become so idealised that they explain very little of the empirical observations of either kind of network. 

In order to address these issues, we propose in this paper to describe a simple model which is both derived directly from computational models built on empirical neuroscience, and also uses information-theoretic results from a relatively new branch of Applied Mathematics, namely the study of coupled Dynamical Systems. We call it the {\sc Feynman Machine} in honour of Richard P. Feynman, for a number of reasons. Feynman was a pioneer in the history of computing, stretching from his work at Los Alamos, to his involvement in early massively parallel algorithms in the 1980's \citep{Hillis:1999:RFC:304763.305699}. Another reason is that he was a colleague of John von Neumann, whose name has become a synonym for the architecture now used in most digital computers, even though von Neumann himself based his designs on concepts derived from then current understanding of cortical neurons \citep{neumann1958computer}. Finally, we hope that Dr. Feynman would appreciate the simplicity of the ideas presented here, ideas which the authors feel Feynman himself might have identified, had he lived to see recent progress in Neuroscience and Applied Mathematics.

We begin by presenting some important results from research into the properties of interacting Dynamical Systems, in particular the Theorem of Floris Takens \citep{takens1981detecting} which proves that models derived from time series signals are essentially true analogues of the system producing the signals. We then present the general Feynman Machine, which is a simple network or hierarchy of regions composed of paired predictive encoder and generative decoder modules. The next section describes how a recent theoretical framework of neural computation can be used to support in neocortex the functionality required by Takens' Theorem and its corollaries, and thus implement a natural Feynman Machine. We then describe a number of novel artificial neural network architectures which also satisfy the requirements, and provide details of software systems we have implemented to examine the properties of the Feynman Machine.

Our artificial Feynman Machines have several interesting properties which distinguish them from existing Deep Learning and similar systems. In particular, due to the much higher density and locality of processing, a Feynman Machine-based system can perform at least comparably while dramatically reducing the footprint in computational power, training data and fine-tuning. Feynman Machines can be arbitrarily split into modules distributed across clusters and the Internet, and systems running on low power devices such as phones can be dynamically and robustly augmented using low-bandwidth connections to larger networks running in the cloud. Models can be trained on powerful infrastructure and transferred for operation and further custom learning on smaller devices. Importantly, the same architecture can be used as a component in unsupervised, semi-supervised, fully supervised and reinforcement learning contexts. A variant - the Routed Predictive Hierarchy - is described, which allows a Feynman Machine to directly control a traditional Deep Learning network by switching it to use spatiotemporally-selected subnetworks.

In summary, the Feynman Machine represents a novel architecture which is both simple enough and powerful enough to explain key structures and function in mammalian neocortex, and also forms a new basis for highly efficient, emergent learning and cognitive processing in machines.

Much of the theoretical basis of this work was originally detailed in \citep{byrne2015symphony}.

The software developed for this work will be available in source code form for non-commercial use at https://github.com/OgmaCorp by September 30, 2016.

\section*{Methods}
\subsection*{A Summary of Relevant Results from Applied Mathematics}
The world of our daily experience is full of things which change over time, either on their own or as a result of our actions. Science is the study of such phenomena, a search for rules and laws which describe their structure and evolution over time, and is based on the premise that the world is lawful and its rules are discoverable. On the other hand, many systems of interest are difficult to model precisely. Some, like the solar system, have too many ``moving parts'' which all influence each other, and we cannot solve their equations exactly, as Poincar\'{e} discovered \citep{poincare1881,poincare1992new} over a century ago. 

A Dynamical System is a mathematical model whose dynamics are characterised by express update rules, typically differential equations 
in continuous systems, or difference equations in discrete time (a comprehensive survey of Dynamical Systems is \citep{strogatz2014nonlinear}).
Study of such systems began with the advent of calculus, and many simple systems have been studied. Simple Harmonic Motion (SHM), 
an idealised approximation of small oscillations of a simple pendulum or spring, is the archetypal dynamical system introduced in
high school physics. Such systems can be solved exactly using calculus, and most systems studied by engineers and scientists are similarly
straightforward to understand and reason about. However, outside the boundaries of approximation used for SHM, even simple springs
are no longer subject to perfect analysis, as their dynamics are nonlinear and can even be chaotic. 

For these reasons, Engineering and Science have historically avoided the problem of analytically insoluble dynamical systems, usually
by approximating the real system by something akin to SHM, a process called linear approximation. With the advent of computers in the mid-20th century, however, researchers have been able to study the dynamics of nonlinear and complex systems, and since the 1960s this has become a 
primary focus of Applied Mathematics. 

In the seminal study which triggered the revolution in Dynamical Systems, 
\citep{lorenz1963} described a simple model of atmospheric convection, using three coupled differential equations. Lorenz' system is an example of {\it deterministic chaos}, which means that the future state of the system can only be estimated confidently over a short period of time, and extending the ``horizon'' requires exponentially more accurate measurements of the current state. This property is shared by a vast number of natural and human systems, a familiar example being the weather, for which tomorrow's forecast might be reliable, while that for seven or more days out will be little more than a reasonable guess.

Floris Takens \citep{takens1981detecting}, basing his study on results going back to \citep{whitney1936differentiable}, proved that
a system such as Lorenz' could be reconstructed in all important details using only a time-series of a single measurement from the
system. A system with manifold dimension $m$ can be reconstructed simply by plotting, for each time $t$, the vector of time-delayed
measurements $\left(x(t),x(t-\tau),x(t-2\tau),...x(t-k\tau)\right)$ of $k>2m+1$ observed values $x$. \citep{Sauer:2006} showed that 
Takens' Theorem holds in less restricted systems, for either $k$ separate measurements at time $t$ or for a set of $k$ time-delayed 
measurements.

The kernel of Takens' Theorem is that the model constructed from the time series is to all intents and purposes {\it the same thing} as the system being observed (formally they are {\it diffeomorphic}). A computer system or brain region which contains the model is capable of using it to produce predictions which have the same properties as those performed on the real system, yet the computer and the brain region need not have any information about the underlying equations of motion which govern the system in the real world. 

A soccer player who runs into the box to head a crossed ball into the net is clearly not solving the simultaneous differential equations of a spinning ball's motion through moving air, under gravity, nor is his run the result of preplanning a sequence of torques generated by his muscles. The player's brain has a network of dynamical systems models which have, through practise and experience, learned to predict the flight of the ball and plan a sequence of motor outputs which will, along with intermediate observational updates and corrections, lead to the desired performance of his skill.

\subsection*{The Feynman Machine Architecture}

A Feynman Machine is a collection of Dynamical Systems modules called {\it regions}, connected together in a network or hierarchy, and to its external world, via sensorimotor channels, each of which carries a (usually high-dimensional) time series signal of some kind. Each region is capable of adaptively learning, representing, predicting and interacting with the spatiotemporal, sensorimotor structure of its ``world''. The internal structure of a region may vary from a single monolithic component to an internal network of components which are each capable of performing part of the task.

Unlike the Turing Machine (or any digital computer), the Feynman Machine is not ``programmed'' in the everyday sense. Instead, the structure of the network, the choice and configuration of regions, and connections to its external world together dictate the functionality and capability of the system, and the actual performance is achieved by online learning of the structure in the world. In natural settings such as neocortex, these hyperparameters are chosen by genetic inheritance and adapted during development. In artificial settings, the setup of the system is the primary design task for the implementor, as illustrated in several examples later in this paper.

In the neocortex, a region corresponds to a multilayer patch of grey matter several millimeters square, sometimes referred to as a cortical macrocolumn or cortical column, several or many such patches being combined to form a ``brain region'' such as V1. In our artificial systems, a region is often synonymous with a layer or level in a hierarchy, and our description uses the term ``layer'' for that reason. 

A Feynman Machine region typically has two ``faces'': the ``visible'', ``downward'' or ``input'' face and the ``hidden'', ``upward'' or ``output'' face. Each face has both inputs and outputs, but their semantics are different for each face. The ``visible inputs'' correspond to sensorimotor inputs to the region, and the ``visible outputs'' to predictions of future inputs, feedback predictive signals to lower Regions and/or control/behavioral/attention signals. The ``hidden outputs'' correspond to encodings of the visible inputs which are directed to visible inputs in higher regions, and the ``hidden inputs'' receive feedback/control/attention/routing signals from higher regions' visible outputs. 

In addition, recurrent internal channels in a region connect internal components. For example, in the Sparse Predictive Hierarchy described later, the hidden output (the predictive encoding) is combined with the hidden input (feedback from above) and fed downward through a decoder to generate a visible output prediction of the next input, and then this is compared with the real next input to generate a prediction error and drive learning. In neocortex, the considerable intralaminar connections serve similar purposes in order to support autonomous region-level learning and modelling.

\subsection*{The Feynman Machine in Neocortex}

While many or most phenomena experienced by animal and human intelligence are governed by underlying physical and mathematical laws, there is no doubt that skilled animal performance does not depend on any explicit knowledge of the hidden differential equations, since only a tiny minority of humans among all mammals are even aware of them, and that is only true since about 1650. Only recently has the field of Applied Mathematics provided a simple and powerful answer to this mystery. An agent with rudimentary processing power can simply ``plot'' the trajectory derived from a time series of measurements, and automatically gain the ability to forecast its actual or contingent future. 

The Feynman Machine is an embodiment of this principle which explains much of the structure of neocortex as found in empirical Neuroscience, and provides a number of principles for building machine intelligence. 

It has been known for over a century that neocortex has a laminar structure which involves very significant local feedback loops, but there is no clear consensus on the reasons for the preservation of this structure over scales of six orders of magnitude across mammalian species, and over the majority of 225 million years of evolution. In addition, it has thus far been a mystery how such an apparently uniform structure can equip various animals with so many apparently disparate skills and capabilities.

We cannot claim to have explained every aspect of computation in neocortex, but this model clearly provides a simple explanation for much of the structure discovered by empirical neuroscience. 

In the arena of small cortical circuits, recent work by Hawkins et al. \citep{hawkins2015neurons} has provided a simple model of the kind of processing which is thought to dominate in single layers of cortex. This model, extended to better cater for temporal structure by using prediction to affect perception, has been detailed in \citep{byrne2015paCLA}. The model, known as the Cortical Learning Algorithm (CLA) of Hierarchical Temporal Memory (HTM), uses a realistic neural model which involves active dendrite segments as local coincidence detectors to register and learn the transitions from one timestep to the next.

\citep{byrne2015symphony} describes an extension of CLA which ascribes roles to all 5 layers (Layers 1, 2/3, 4, 5 and 6) of neocortex in a region. In terms of the Feynman Machine, the main role of Layer 4 is to act as the downward input face, receiving afferent sensorimotor inputs and predicting individual transitions; Layer 2/3 temporally pools over these transitions and acts as part of the encoder's hidden output (it also receives feedback from higher regions); Layer 5 combines inputs from Layer 2/3 and from higher regions via Layer 1, thus forming the main hidden input face; and both Layer 5 and 6 form outputs: Layer 5 to motor areas as well as upwards, and Layer 6 produces attention/feedback/modulation signals to lower regions and Layer 4.

While this is clearly more complex than the clean model of the Feynman Machine we use in computers, there is a definite homology between the primary processing pathways in neocortex and our design. The ``feedforward'' part of a region, formed mainly by Layers 4 and 2/3, represents and predicts the evolution of the sensory inputs, playing the role of the Encoder module in our design. The ``feedback'' part, formed mainly by Layers 5 and 6, produces feedback and behaviour signals and modulation of lower regions, playing the role of the Decoder module in the Feynman Machine architecture. Internal connections, primarily between Layers 6 and 4 and between Layers 2/3 and 5, play the roles of the lateral and prediction error connections in our design.

We show in our experiments that the Feynman Machine operation is not very sensitive to the exact choice of algorithm in each Encoder and Decoder module, and that different connection schemes may be used to connect modules internally as well as between regions. The important thing is that each region has an internal feedback loop where encodings affect decodings and vice versa. While the Neuroscience evidence of strong intralaminar connectivity conforms exactly with this scheme, the precise wiring of the individual axons and their semantic content and purpose is unknown. Our model provides candidates for the primary role of these connections and predictive constraints for the wiring and effects of their signals.

The model of the brain as a distributed Feynman Machine network has strong explanatory power. The network diagram or connectome of mammal brains varies considerably from one animal group or species to another, and there is also genetically-determined variation between individual regions in a singe brain, reflecting the various specific roles of each region in the organism's performance. Cadieu et al \citep{cadieu2014deep} contains strong evidence that the fast feedforward recognition pathway in visual neocortex resembles a convolutional neural network, but there is no artificial neural network which models the longer-timescale processing of visual information involving local and inter-region feedback loops. The Feynman Machine is the first such model which is simple enough to test by simulation, as well as powerful enough to explain the flexible, general purpose processing power of large networks of communicating regions.

\subsection*{Artificial Feynman Machines}

Evolution is often limited by its inability to escape local maxima in the fitness landscape. Designers of artificial systems can jump that barrier and devise solutions which dramatically surpass those found in nature, particularly in terms of simplicity and measurability. Our software implementations of the Feynman Machine share a common architecture: a ladder-like hierarchy of paired encoders and decoders. This resembles the ladder networks of \citep{rasmus2015semi}, but extends it to perform predictions of the next timestep of data.

\begin{figure}[ht]
 \centering
 \includegraphics{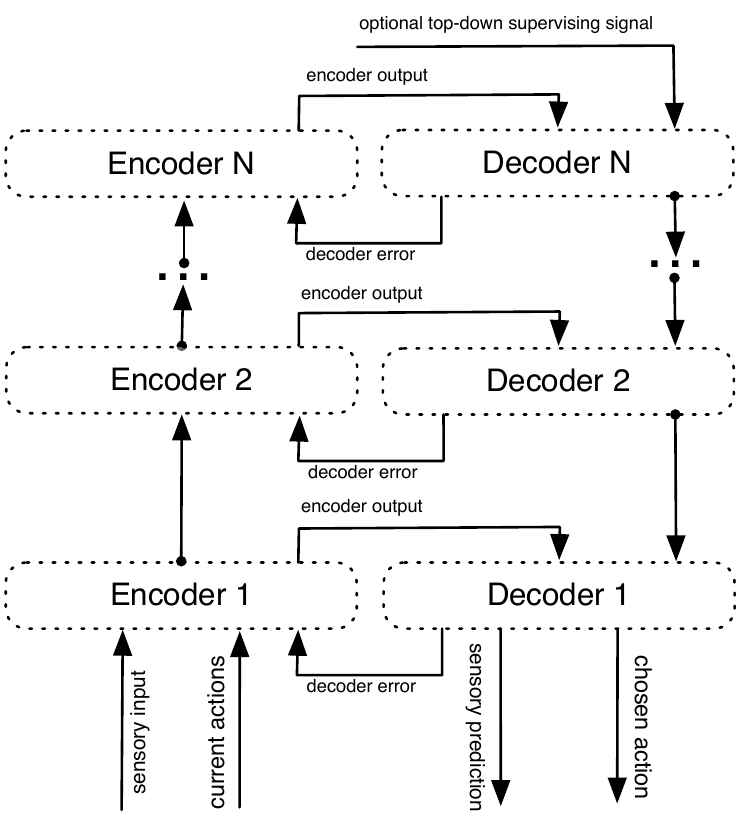}
 \caption{The Sparse Predictive Hierarchy Feynman Machine.}
 \label{fig:fig1}
\end{figure}

Information flows both up the hierarchy (feature extraction), and down the hierarchy (prediction). Each encoder-decoder pair or {\it predictor} attempts to predict its inputs, ie the state of the layer below it, one timestep in advance. Lower layers use feedback information from the layers above to improve their own predictions.

Each encoder/decoder pair is a kind of spatiotemporal autoencoder, receiving information, mapping to a hidden representation, and using that to predict the next frame of information. Predictions are compared with the actual next input, and this prediction error can be used to drive learning.

Because of the generality of the theory of networked Dynamical Systems, we can choose to use many designs for the encoders and decoders, based on their different properties. The design we've found most useful in experiments and application is an online, spatiotemporal modification of the $k$-sparse autoencoder \citep{makhzani2013k}.

\subsection*{Spatiotemporal $k$-Sparse Autoencoder}

The hidden representation is sparse: most units are zero at any given instant. We use $k$-sparsity, where the $k$ most active units are kept as is and the rest are set to zero, mimicking lateral inhibition found in biological systems. Forming a sparse distributed representation (SDR) forces the encoder to learn more orthogonal weight vectors, which avoids catastrophic interference. To ensure that weights are also sparse and well-distributed in the time dimension, we impose temporal sparsity constraints.


\subsubsection*{Description}
We will describe the encoder in terms of dense connectivity, although in our experiments we used sparse connectivity patterns (local connections). A pseudocode version of the Sparse Predictive Hierarchy is presented in the Supplementary Material.

\paragraph{Derived Inputs} In a simple one-shot memory mechanism, the {\it derived inputs} $\mathbf{d}$ to the encoder are computed from the previous derived inputs $\bar{\mathbf{d}}$ and the current inputs $\mathbf{x}$:

\begin{align}\label{eq:input}
\mathbf{d}^1 &= \mathbf{x}\\
\mathbf{d}^2 &= \lambda\bar{\mathbf{d}}^2 + (1-\lambda)\mathbf{x}
\end{align}

where $\mathbf{d}^1$ is the instantaneous value of the visible unit input, and $\mathbf{d}^2$ is a running average of the inputs. $\lambda$ is the one-shot memory decay multiplier.

\paragraph{Stimulus} The {\it stimulus} $\mathbf{s}$ each hidden unit receives from the visible units $\mathbf{d}$ is a simple linear combination:

\begin{align}\label{eq:stimulus_full}
\mathbf{s}^1 &= \bar{\mathbf{s}}^1 +
          \mathbf{W}_{enc}(\mathbf{d}^1 - \mathbf{d}^2)\\
          \label{eq:stimulus_full_weights}
\mathbf{s}^2 &= \bar{\mathbf{s}}^2 + \|\mathbf{W}_{enc}\|^2_{row}
\end{align}

Where $\mathbf{W}_{enc}$ is a matrix of weights with a number of rows equal to the number of hidden units and a number of columns equal to the number of visible units. The norm in (\ref{eq:stimulus_full_weights}) is that of the row of weights for each stimulus value.

In practice, we do not use fully-connected weights, but instead sum weighted inputs to the encoder over a window of radius $r$ around the projected position $(k_0, l_0)$ in $\mathbf{d}$ as follows:

\begin{align}\label{eq:stimulus}
(k_0, l_0) &= projection_{S\rightarrow D}(i,j)\\
s_{ij}^1 &= \bar{s}_{ij}^1 +
          \sum_{kl \in \mathcal{N}(k_0, l_0,r)}{
            (d_{kl}^1 - d_{kl}^2)\mathbf{w}_{ijkl} }\\
s_{ij}^2 &= \bar{s}_{ij}^2 +
          \sum_{kl \in \mathcal{N}(k_0, l_0,r)}
            {(\mathbf{w}_{ijkl})^2}
\end{align}

This more involved calculation for practical 2D inputs and layers (and 4D locally connected matrices) will not be repeated in this section. See {\it \nameref{S1_Algorithms}} for the complete pseudocode for all calculations.

Note that $\mathbf{s}^2$ is a "plane" of values representing the running average squared norm of the weights used to compute $\mathbf{s}^1$. These values are used to scale memory in encoder weight learning (see below).

\paragraph{Activation} The activation $\mathbf{a}$ for the hidden units is:

\begin{align}\label{eq:activation}
\mathbf{a} &= \beta \bar{\mathbf{a}} + (1-\beta)(\mathbf{s}^1 + \bar{\mathbf{b}})
\end{align}

$\mathbf{b}$ represents the bias for the hidden units. $\beta \in [0,1]$ is the temporal pooling decay factor.



\paragraph{$k$-Sparsity} In a fully-connected, non-local setup, build a heap around the units, and take the top $k$ elements to be the active units:

\begin{equation}\label{eq:inhibition_full}
\mathbf{z}_j = \mathbf{1}(j \in supp_k(\mathbf{a}))
\end{equation}

where $supp_k(\mathbf{a})$ is the set of indices $j$ of the top $k$ elements of the activations $\mathbf{a}$.

In our software, $k$-Sparsity is implemented locally by counting the number of activations $n^*_{ij}$ higher than the current one, and setting the unit to zero if this exceeds the threshold target sparsity $\rho$:

\begin{align}\label{eq:inhibition}
\forall i, j\ \ n^*_{ij} &= \sum_{kl \in \mathcal{N}(ij,r)}
        {\mathbf{1}(a_{kl} \geq a_{ij})}\\
z_{ij} &= \mathbf{1}(n^*_{ij} < \rho |\mathcal{N}(ij, r)|)
\end{align}







\paragraph{Decoding} The encoder thus produces a $k$-sparse hidden output state $\mathbf{z}$. To train the system, we want the decoder errors to influence the encoding, so the encoder and decoder are closely coupled. The decoder computes its prediction $\mathbf{y}_l$ using two sources: the feedback input $\mathbf{z}_{fb} = \mathbf{y}_{l+1}$ coming from a higher layer decoder's prediction (or the hidden states $\mathbf{z}_l$ in the case of the top layer), and the lateral input $\mathbf{z}_{lat} = \mathbf{z}_l$ coming from the paired encoder hidden states.

The fully-connected decoding is simple:

\begin{align}\label{eq:prediction}
\mathbf{y}_{fb} &= \mathbf{W}_{fb}\mathbf{z}_{fb}\\
\mathbf{y}_{lat} &= \mathbf{W}_{lat}\mathbf{z}_{lat}
\end{align}

The locally-connected case is calculated similarly to the encoding stage (uses a window of radius $r$). The local version can be simulated by setting $\mathbf{W}_{ijkl} = 0$ for all $(k,l)$ outside the window of radius $r$ around $(k_0, l_0) = projection_{\mathbf{y}\rightarrow\mathbf{z}}(i,j)$.

Now combine the feedback and lateral predictions to produce the decoder output $\mathbf{y}$ using a blending factor $b$:

\begin{align}\label{eq:combine_prediction}
\mathbf{y} &= b \mathbf{y}_{fb} + (1-b)\mathbf{y}_{lat}
\end{align}



\paragraph{Prediction Error} This describes a full activation cycle of the encoder-decoder pair. To train it, we use a local backpropagation algorithm. First, compute the prediction error $\mathbf{\epsilon}_{y}$:

\begin{align}\label{eq:prediction_error}
\mathbf{\epsilon}_{y} &= \mathbf{x}_t - \mathbf{y}_{t-1}\ \ \ \textrm{// bottom layer}\\
\mathbf{\epsilon}_{y} &= \mathbf{d}_t - \mathbf{y}_{t-1}\ \ \ \textrm{// upper layers}
\end{align}

Backpropagate the error to the hidden units:

\begin{align}\label{eq:hidden_error}
\mathbf{\epsilon}_{z} &= \mathbf{W}^T_{lat} \mathbf{\epsilon}_{y}
\end{align}

Update the decoder weight matrices:

\begin{align}\label{eq:decoder_weight_update}
\Delta \mathbf{W}_{fb} &= \alpha_{fb} (\mathbf{\epsilon}_{y} \otimes \bar{\mathbf{y}}_{fb})\\
\Delta \mathbf{W}_{lat} &= \alpha_{lat} (\mathbf{\epsilon}_{y} \otimes \bar{\mathbf{y}}_{lat})
\end{align}


and the encoder weight matrices:

\begin{align}\label{eq:encoder_weight_update}
\mathbf{W}_{enc} &= \mathbf{\kappa} \cdot \bar{\mathbf{W}}_{enc} + \alpha_{enc} (\mathbf{\epsilon}_{z} \otimes (\bar{\mathbf{d}}^1 - \bar{\mathbf{d}}^2))
\end{align}

where $\alpha$ are learning rates, $\otimes$ is the outer product of vectors, and $\mathbf{\kappa}$ is a scaling forgetting factor vector which is derived from the stimulus weight norms $\mathbf{s}^2$ using:

\begin{align}\label{eq:scale_weight_update}
\mathbf{\kappa}_j = \frac{1}{\sqrt{max(10^{-4}, \mathbf{s}^2_{j})}}
\end{align}

The old matrix $\bar{\mathbf{W}}_{enc}$ is multiplied columnwise by $\mathbf{\kappa}$.

This learning rule is akin to Spike-time Dependent Plasticity (STDP), since it strengthens weights which connect inputs and closely following output spikes, and weakens those which connect temporally anticorrelated outputs.

The bias vector $b$ is updated such that it helps maintain the lifetime sparsity of a unit (to prevent dead units):

\begin{equation}\label{eq:bias_weight_update}
\Delta \mathbf{b} = \gamma (\rho \mathbf{1} - \mathbf{z})
\end{equation}

where $\rho$ is the ratio of active units to inactive units ($k$ divided by the number of hidden units), and $\gamma$ is another learning rate.

This resulting spatiotemporal autoencoder handles credit assignment using the two mechanisms: a simple one-shot memory mechanism, and recurrent inputs (as part of $\mathbf{d}$). This autoencoder can be trained online without multiple iterations per timestep while needing minimal computational resources.

\section*{Predictive Hierarchies}

\subsection*{Approaches}
Feeding hidden unit states as input to the next encoder/decoder pair results in a hierarchical spatiotemporal feature extractor as information is passed upwards. For the downwards pass, there are generally two methods of forming predictions:

\begin{enumerate}
    \item Top-down feedback connections/reconstruction
    \item Modulation of another (top-down) hierarchy by hidden states
\end{enumerate}

For the former, we simply need to add several perceptrons (one for each layer) where each perceptron predicts the input of the associated encoder/decoder pair from both the hidden units and feedback information. The feedback is simply the decoder prediction from the next higher layer. As a result, one can start at the top of the hierarchy, and predict one layer at a time taking the feedback information from the previous layer into account. This is the approach described above, which we call the {\it Sparse Predictive Hierarchy} (SPH, Figure \ref{fig:fig1}).

\begin{figure}[ht]
 \centering
 \includegraphics{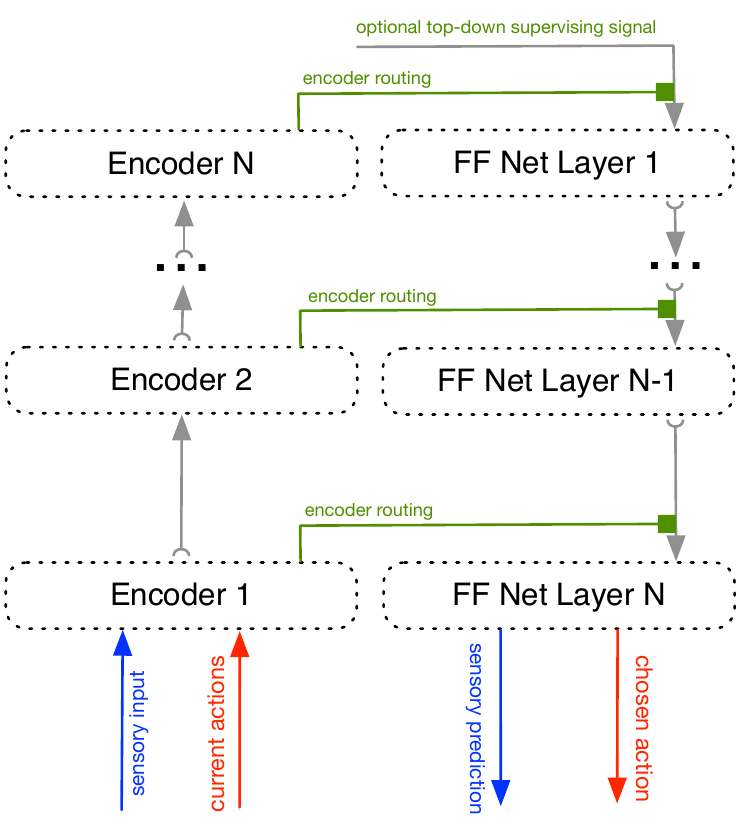}
 \caption{The Routed Predictive Hierarchy Feynman Machine.}
 \label{fig:fig2}
\end{figure}

Alternatively, the downpass can some other graphical model that specializes in predictions, where the sparse states from the encoder/decoder pairs are used to modulate another hierarchy (which could be both top-down or bottom-up). Only the portions of this modulated network where the corresponding encoder hidden states are non-zero are used for computation (activation and learning). We call this approach {\it network routing}, as it essentially routes the connectivity patterns of another network. The entire hybrid network is termed a {\it Routed Predictive Hierarchy} (RPH, Figure \ref{fig:fig2}).

If a traditional (inverted) feedforward network is used as the routed network, the key difference between SPH and RPH is that in the former, the feedback inputs are added to the lateral inputs, whereas in the routed hierarchy they are multiplied.

\paragraph{Reinforcement Learning}

Much recent progress in Deep Learning has been concentrated in regimes where large amounts of static labelled data is available for supervised training. However, leaders in the field point to the limitations of this approach and call for methods which use unsupervised learning, and ideally reinforcement learning, as found in mammalian intelligence.

Our Sparse Predictive Hierarchy (SPH), as described above, is fully capable of online, unsupervised learning and prediction of streaming, high-dimensional data. In order to utilise the SPH in a reinforcement context, we have two primary choices, corresponding to the two network approaches mentioned above.

The first approach is to construct a reinforcement learning agent which uses an embedded SPH as a producer of sensorimotor predictions. The input to the SPH is a combined vector $(\mathbf{x}|\mathbf{a})$, where $\mathbf{x}$ is the sensory input, and $\mathbf{a}$ is a vector representing the current action. The output of the bottom-level decoder is then a prediction-action vector $(\mathbf{y}|\mathbf{\hat{a}})$ in which $\mathbf{y}$ is the sensory prediction and $\mathbf{\hat{a}}$ is a vector representing the network's chosen action.

We have tested this design extensively on several RL tasks, both in-house and publicly available (in particular a subset of the OpenAI Gym tasks \citep{openai}) and achieve results comparable with leading methods, but using far less computational power.

The second approach uses the encoder ladder to route the activity of a second network. In this case the encoder states control a population of very simple RL agents, thus recruiting different subpopulations of learner-actors depending on the spatiotemporal context. This approach is in some cases more successful than the embedded approach, and is currently being evaluated on the difficult set of ALE video game tasks. We are testing various encoder designs on the hybrid architecture, each of which has different tradeoffs in terms of long-term stability, weight norm control, and computational cost. Our current leading options appear to be a) a variant on the encoder described above, with refractory traces to inhibit persisting SDRs, and b) a SAILnet-like iterative solver which performs a cycle of spiking and inhibition for each input timestep (of the order of tens of cycles per input).

\section*{Results}

\subsection*{Implementations}

Implementations of the architectures detailed in this paper have been developed in several languages, on both CPU and CPU/GPU platforms, and will be available in source code form for non-commercial use at https://github.com/OgmaCorp from September 30, 2016.

The primary implementation for both experimental and production use is {\sc OgmaNeo}, a C++ library that uses OpenCL 1.2 and runs on Windows and Linux PCs, as well as modern Apple Macintosh computers with NVidia GPUs. Future versions of this software will also use NVidia's CUDA directly. 

OgmaNeo is provided as a library and is accompanied by a suite of software, including a visual network/hierarchy editor/builder, a set of Python bindings, and a number of experiments and demonstration programs in C++ and Python which exercise the Feynman Machine in various ways.

Python package {\sc PyOgmaNeo} contains bindings for the main OgmaNeo library, allowing for rapid experimentation on algorithms.

\subsubsection*{Implementations in Progress (Fall 2016)}

{\sc GoOgmaNeo} is an experimental implementation in Google's Go language, aimed at exploring parallel, cross-platform designs. {\sc JOgmaNeo} is an implementation written in Java, and {\sc P3OgmaNeo} is a client/visualiser in Processing which uses it as a back end. Finally, {\sc EOgmaNeo} is an event-based implementation written in C++ for the low-power Raspberry Pi.

\subsection*{Experimental and Development Contexts}
\label{Contexts}

\subsubsection*{Video Prediction}
The Sparse Predictive Hierarchy has been adapted for frame-to-frame video prediction. Learning is a straightforward matter of generating prediction errors by comparing the output prediction $\mathbf{y}_{t-1}$ with the true next frame $\mathbf{x}_t$.

When provided with the first few frames ($\tau_{init}$) of a learned sequence, the system proceeds to play out a learned version of the same sequence, driven by feeding the prediction frame $\mathbf{y}_{t-1}$ as the new input $\mathbf{x}_t$, for $t > \tau_{init}$

For this example, we used a four layer hierarchy with square layers of sizes 128, 96, 64, 32 (from bottom to top). Training took approximately 20 seconds with 16 passes over the video. The source video is 47 frames long. We used an AMD R290 GPU to produce these results.

\begin{figure}[ht]
 \centering
 \includegraphics[scale=0.58]{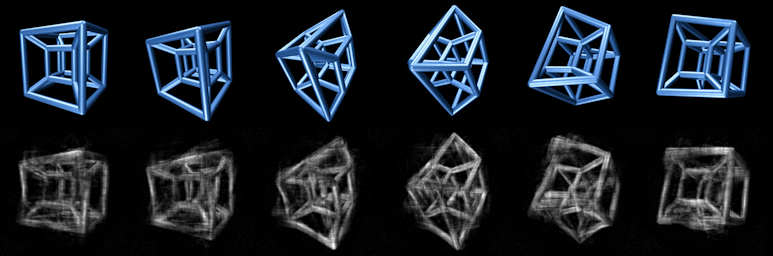}
 \caption{Video Prediction: Sample frames of a synthetic video sequence (top) and the corresponding self-generated output frames (bottom) from a Sparse Predictive Hierarchy.}
 \label{fig:fig3}
\end{figure}

[The system also performs well on natural scenes. We are preparing examples using public domain video for future versions of this paper]

\subsubsection*{Video Super-resolution} Most state-of-the-art methods use only single frames to perform video super-resolution, or deliberately employ shuffling of frames to make successive images as i.i.d. (independent, identically distributed) as possible. Only a few methods exploit the temporal coherence of real-world video sequences. Our methods are designed to maximally exploit the spatiotemporal coherence of objects in real-world video.

The basic method of training is to provide a low-resolution ($\hat{m} \times \hat{n}$ pixels) video stream $\mathbf{\hat{x}}_t$ to the lowest layer of the Sparse Predictive Hierarchy, and have the decoder output use a size equivalent to the full resolution video ($m \times n$). The output decoding $\mathbf{y}$ of the bottom layer is then compared with the high-resolution training input $\mathbf{x}$ in order to generate the prediction errors for learning.

[We are preparing examples using public domain video for future versions of this paper]

\subsubsection*{Noise Reduction} We have evaluated using sparse predictive hierarchies within the areas of audio and image noise reduction.

\paragraph{Image noise reduction} Image denoising applications are trained by providing as input a noised input image $\mathbf{\dot{x}}$ and generating prediction errors by comparing the decoder output $\mathbf{y}$ with the original clean image $\mathbf{x}$.

\paragraph{Audio noise reduction} This is typically performed within a frequency domain \citep{yu2008audio, siedenburg2012audio, stowell2015denoising}. Using a modified Wigner distribution function, discrete wavelet transform, or short-time Fourier analysis (with appropriate window function such as Hann or Blackman-Harris, for example). Potentially using a finite impulse response Wiener–Kolmogorov filter or spectral subtraction methods. Frequency domain analysis can also be affected when considering reversibility \citep{decorsiere2015inversion}. For example, when dropping or reusing the phase contribution. Techniques exist to approximate and/or reduce this loss during inversion \citep{zhu2007real}.

With the online-learning aspect of a sparse predictive hierarchy, we have experimented with single-channel denoising of a signal while remaining in the time domain. This is similar to our work using predictive hierarchies for frame-frame video prediction. Following other audio denoising techniques we have used objective difference grade scores from the ITU-R recommendation BS.1387-1 perceptual evaluation of audio quality \citep{thiede2000peaq}, to evaluate the performance of predictive hierarchies.

\subsubsection*{Anomaly Detection}
The Sparse Predictive Hierarchy is well-suited to online anomaly detection, since it has the advantageous properties of Hierarchical Temporal Memory systems (but without the need for type-specific encoders, and not limited to one layer, as HTM is). We have applied a variant of the SPH on the Numenta Anomaly Benchmark \citep{lavin2015evaluating}, achieving results comparable with HTM, but with a dramatically lower computational cost.

Applications of anomaly detection include stroke prediction, cardiac monitoring, clinical monitoring in intensive care settings, monitoring of dementia sufferers, security, fraud detection, monitoring of industrial plant and equipment, energy production, vehicle maintenance, counterterrorism, etc.

\paragraph{Reinforcement Learning}

As described above, we have tested both the embedded Sparse Predictive Hierarchy and the hybrid Routed Predictive Hierarchy controlling swarms of simple RL agents, on both internally developed and publicly available tasks. Initial results are more than encouraging, and we anticipate that achieving RL performance matching and exceeding the state of the art will be our focus in the near future.

Applications of the RL frameworks include self-driving cars, robotics, remote sensing, autonomous drones, robotic engineers (e.g for space exploration, hazardous repairs, etc), and many more.

\section*{Conclusion}

We have presented a novel neural network architecture, the {\sc Feynman Machine}, which derives its predictive power from the intrinsic modelling within coupled, adaptive Dynamical Systems. We have identified a homologue of the Feynman Machine in mesoscale neocortical circuits, and proposed this as a new core functional component of mammalian intelligence.

A number of software implementations of the Feynman Machine have been developed and we have been investigating their computational properties in a number of test domains. We're confident that this avenue of research shows significant promise in advancing the state of the art in machine intelligence. In particular, the architecture forms a bridge between systems based on hard Neuroscience and more ``brain-inspired'' Deep Learning techniques, to the benefit of both approaches.

A recent paper \citep{carmantini2016modular} discusses the computational properties of networks of idealised Dynamical Automata. Examining the potential connections of these theoretical results with our more empirical developments would seem worthwhile for future work.

\section*{Supporting Information}

\subsection*{S1 Sparse Predictive Hierarchy Algorithms}
\label{S1_Algorithms}
Each predictive hierarchy is constructed as follows (for $l$ layers, $m_{in} \times n_{in}$ inputs, and $m_{out} \times n_{out}$ outputs):
\begin{enumerate}
  \item allocate sparse predictors for each layer
  \item allocate 2D arrays for each layer for pooling data. array size is proportional to the layer size.
\end{enumerate}

Each sparse predictor contains a collection of visible and hidden layers indexed by $i$. Visible (input-facing) layers have size $m^i_{vis} \times n^i_{vis}$, and hidden (output-facing) layers have size $m^i_{hid} \times n^i_{hid}$.

Each sparse predictor visible layer is constructed as follows:
\begin{enumerate}
  \item allocate $m^i_{vis} \times n^i_{vis}$ 2D arrays for the derived input $D$ and reconstruction error $E_Y$, zero filled.
  \item if the layer is designed for input; allocate 4D arrays $\mathbf{W}_{enc}$ for encoder weights, initially random. In most cases, connections are sparse, so weight matrices are of size $\hat{m}^i_{vis} \times \hat{n}^i_{vis} \times m^i_{hid} \times n^i_{hid}$ for a chosen receptive field (window) size $\hat{m}^i_{vis} \times \hat{n}^i_{vis}$. In other implementationsweight matrices are fully connected, so the dimensions are $m^i_{vis} \times n^i_{vis} \times m^i_{hid} \times n^i_{hid}$.
  \item if the layer is designed to perform predictions; allocate 4D arrays for feedback decoder weights $\mathbf{W}_{fb}$ and lateral decoder weights $\mathbf{W}_{lat}$, and 2D arrays for predictions $Y$, feedback predictions $Y_{fb}$, and laterally-sourced predictions $Y_{lat}$. 2D arrays are of size $m^i_{vis} \times n^i_{vis}$ and initialised to zero, 4D weight matrices are sparse or fully connected, depending on the design.  Weight arrays initially random in a given range.
\end{enumerate}

Each sparse predictor hidden layer is constructed as follows:
\begin{enumerate}
  \item allocate $m^i_{hid} \times n^i_{hid}$ 2D arrays for activations $A$, biases $B$, hidden states $Z$, stimulus summations $S$ and error summations $E_Z$, next layer inputs $Z^{max}$ (if max-pooling is enabled).
  \item randomly initialize the biases array $B$, and zero fill the activation and state arrays.
\end{enumerate}

Several of the 2D arrays and 4D matrices (including $D$, $S$, $Z$ in some versions, and the $\mathbf{W}$'s) described above will actually contain a second ``plane'' of values, often used to store some history information of additional utility when combined with the values in the first plane. For example, the derived input $D$ stores a copy of the immediate input $x_{ij}$ in $d^1_{ij}$, along with a moving average of past inputs in $d^2_{ij}$, and the temporal difference $d^1_{ij} - d^2_{ij}$ is used in several places in the system.

Since the hierarchy is designed to model temporal evolution of the data, each sparse predictor is constructed with a "previous" copy of each 2D array and 4D weight matrix described above, denoted by a bar symbol, thus for example the previous activations are denoted $\bar{A}$. At the beginning of each timestep, the arrays and weight matrices are swapped, so for example $\bar{A}$ takes on the values of $A$, and $A$ is either reset to zero or simply overwritten with new values during the processing.

Depending on the platform, hardware or programming system employed in a particular implementation, it may be more appropriate to employ 3D matrices in place of 4D matrices as described in this section. In such cases the dimensions of the 3D matrix would be $(\hat{m}_{vis} \times \hat{n}_{vis}, m_{hid}, n_{hid})$ and the element $\mathbf{w}_{ijkl}$ of the 4D matrix would be the element $\mathbf{w}_{\hat{i}kl}$ of the 3D matrix, where $\hat{i} = i \hat{n}_{vis} + j$. Similar reshaping to use only 1D arrays if necessary is also feasible using similar formulae.

In the following, values may be calculated by ranging over a neighbourhood $\mathcal{N}(ij,r)$ of some position $(i,j)$ with a "radius" $r$. This neighbourhood ranges from $i - r$ to $i + r$ in one dimension, and from $j - r$ to $j + r$ in the other. In some designs the "radii" in each dimension may not be identical, so in general $r = (r_x, r_y)$ may define a rectangular neighbourhood or window which is used for the calculation. In some cases where the sizes of the arrays differ, the system calculates a "center" $(k_0, l_0)$ in the second array corresponding to the position $(i, j)$ in the first, and this is denoted as follows $(k_0, l_0) \longleftarrow projection_{S\rightarrow D}(i,j)$. In this case the neighbourhood will be $\mathcal{N}(k_0,l_0,r)$. Finally, depending on the design or parameters chosen by the user, the window may or may not include the "center" element itself (eg inhibition ignores the center).

\subsubsection*{Predictive Hierarchy processing}

\paragraph{simStep(inputs, learn)} runs the hierarchy encoding and decoding passes, generates predictions and performs learning optionally.

\begin{procedure}[H]
\DontPrintSemicolon
\KwData{image $inputs$, learning control $learn$}
\KwResult{}
\Begin{
    $encodeHierarchy(inputs)$\;
    $decodeHierarchy(inputs)$\;
    $hierarchyErrors(inputs)$\;
    \lIf{learn?}{ $learnHierarchy(inputs)$ }
}
\caption{simStep($inputs$, $learn?$)}
\end{procedure}

\paragraph{encodeHierarchy()} pass input into the hierarchy, and run the up-pass for each layer.

\begin{procedure}[H]
\DontPrintSemicolon
\KwData{image $inputs$}
\KwResult{}
\Begin{
    \lIf{whiten}{
       $inputs \longleftarrow whiten(inputs)$
    }
    $visibles \longleftarrow inputs$\;
    \ForEach{layer $l$}{
        $l.states \longleftarrow l.activateEncoder(visibles, \lambda)$\;
        $l.maxPool \longleftarrow maxPoolKernel(l.states, poolRadius)$\;
        $visibles \longleftarrow l.maxPool$\;
    }
}
\caption{encodeHierarchy($inputs$)}
\end{procedure}

\paragraph{activateEncoder() and its kernels} pass the hierarchy's inputs up from layer to layer.

\begin{procedure}[H]
\DontPrintSemicolon
\KwData{visible input $X$, previous $\bar{X}$, biases $B$, weights $\mathbf{W}_{enc}$, decay $\lambda$, pooling $\beta_{pool}$}
\KwResult{derived input $D$, stimulus $S$, activations $A$, hidden states $Z$}
\Begin{
    $D \longleftarrow deriveInputKernel(X, \bar{D}, \lambda)$\Comment*[r]{derived input}
    \If{layer used for input}{
        $S \longleftarrow encodeKernel(D, \bar{S}, \mathbf{W}_{enc}, r)$\Comment*[r]{compute stimulus}
    }
    $A \longleftarrow activateKernel(S, \bar{B}, \bar{A}, \beta_{pool})$\Comment*[r]{compute activations}
    $Z \longleftarrow solveHiddenKernel(A, r_{inhibit})$\Comment*[r]{compute hidden states}
}
\caption{activateEncoder($visibles$, $\lambda$)}
\end{procedure}

\paragraph{deriveInputKernel()} produces a pair of inputs to the encoder, one a copy of the input and the other a moving average of recent inputs.

\begin{procedure}[H]
\DontPrintSemicolon
\KwData{visible inputs $X$, previous output $\bar{D}$, decay $\lambda$}
\KwResult{derived output $D$, $d_{ij}^1$ is the input copied, $d_{ij}^2$ is decayed}
\For{$x_{ij} \in X$}{
    $d_{ij}^1 \longleftarrow x_{ij}$\;
    $d_{ij}^2 \longleftarrow \lambda\bar{d}_{ij}^2 + (1-\lambda)x_{ij}$\Comment*[r]{decay previous}
}
\caption{deriveInputKernel($inputs$, $outputsPrev$, $\lambda$)}
\end{procedure}

\paragraph{encodeKernel()} produces the encoder's stimulus, which is a weighted sum of each unit's derived inputs. The second element of the stimulus is a running sum of the squared norm of the weights used for each unit.

\begin{procedure}[H]
\DontPrintSemicolon
\KwData{derived inputs $D$, previous stimulus $\bar{S}$, weights $\mathbf{W}_{enc}$, window radius $r$}
\KwResult{new stimulus $S$}
\For{$\bar{s}_{ij} \in \bar{S}$}{
        $(k_0, l_0) \longleftarrow projection_{S\rightarrow D}(i,j)$\;
        $s_{ij}^1 \longleftarrow \bar{s}_{ij}^1 +
          \sum_{kl \in \mathcal{N}(k_0, l_0,r)}{
            (d_{kl}^1 - d_{kl}^2)\mathbf{w}_{ijkl} }$\;
        $s_{ij}^2 \longleftarrow \bar{s}_{ij}^2 +
          \sum_{kl \in \mathcal{N}(k_0, l_0,r)}
            {(\mathbf{w}_{ijkl})^2}$
    }
\caption{encodeKernel($derivedInput$, $prevStimulus$, $weights$, $radius$)}
\end{procedure}

\paragraph{activateKernel()} updates the hidden activations by combining the stimuli and biases.

\begin{procedure}[H]
\DontPrintSemicolon
\KwData{stimulus $S$, existing biases $\bar{B}$, previous activations $\bar{A}$, pooling $\beta_{pool}$}
\KwResult{new activations $A$}
\For{$\bar{a}_{ij} \in \bar{A}$}{
        $a_{ij} \longleftarrow \beta \bar{a}_{ij} + (1-\beta)(s^1_{ij} + \bar{b}_{ij})$\;
    }
\caption{activateKernel($stimulus$, $\bar{biases}$, $\bar{activations}$, $pooling$)}
\end{procedure}

\paragraph{solveHiddenKernel()} implements local inhibition on the activations to produce a sparse binary hidden state.

\begin{procedure}[H]
\DontPrintSemicolon
\KwData{activations $A$, radius $r$}
\KwResult{new hidden states $Z$}
\For{${a}_{ij} \in A$}{
    $inhibition \longleftarrow \sum_{kl \in \mathcal{N}(ij,r)}
        {\mathbf{1}(a_{kl} \geq a_{ij})}$\;
    $z_{ij} \longleftarrow \mathbf{1}(inhibition < sparsity \times \|\mathcal{N}(ij, r)\|)$\Comment*[r]{fire if winner}
}
\caption{solveHiddenKernel($activations$, $radius$)}
\end{procedure}

\paragraph{maxPoolKernel()} creates a summarised representation of the hidden state.

\begin{procedure}[H]
\DontPrintSemicolon
\KwData{hidden states $Z$, radius $r$}
\KwResult{max-pooled $Z^{max}$}
\For{$z^{max}_{ij} \in projection_{Z \rightarrow Z^{max}}(Z)$}{
    $z^{max}_{ij} \longleftarrow \max_{kl \in \mathcal{N}(ij,r)} z_{kl}$\;
}
\caption{maxPoolKernel($states$, $radius$)}
\end{procedure}

\paragraph{Decoding} information passes down the hierarchy, as each layer generates predictions by decoding its encoder's output, combined with feedback from higher layer decoder predictions.

\begin{procedure}[H]
\DontPrintSemicolon
\KwResult{prediction $Y_l$ for each layer}
 \For{each layer $l$ in reverse order}{
        \lIf{not top layer}{
            $Z_{fb} \longleftarrow Y_{l+1}$
        }
        \lElse{
            $Z_{fb} \longleftarrow Z$
        }
        \If{$l$ predicts}{
            $Y_{fb} \longleftarrow decodeKernel(Z_{fb}, \mathbf{W}_{fb}, r_{fb})$\;
            $Y_{lat} \longleftarrow decodeKernel(Z, \mathbf{W}_{lat}, r_{lat})$\;
            $Y_l \longleftarrow combinePredKernel(Y_{fb}, Y_{lat}, blend)$\;
        }
}
\caption{decodeHierarchy()}
\end{procedure}

\paragraph{decodeKernel()} transforms either the hidden or feedback inputs to the decoder using a weight matrix, forming a partial prediction.

\begin{procedure}[H]
\DontPrintSemicolon
\KwData{source $Z$, weights $\mathbf{W}$, window radius $r$}
\KwResult{predictions $Y$}
\For{$y_{ij} \in Y$}{
        $(k_0, l_0) \longleftarrow projection_{Y\rightarrow Z}(i,j)$\;
        $y_{ij} \longleftarrow \sum_{kl \in \mathcal{N}(k_0,l_0,r)}{
            z_{kl} \mathbf{w}_{ijkl} }$\;
    }
\caption{decodeKernel($source$, $weights$, $radius$)}
\end{procedure}

\paragraph{combinePredKernel()} blends the feedback and lateral predictions to produce a continuous (typically for the output decoder) or optionally locally sparse prediction output from the decoder.

\begin{procedure}[H]
\DontPrintSemicolon
\KwData{feedback and lateral predictions $Y^{fb}$ and $Y^{lat}$, blend ratio $b$, radius $r$}
\KwResult{predictions $Y$}
\For{$y_{ij} \in Y$}{
    $y_{ij} \longleftarrow (1 - b)y^{fb}_{ij} + b y^{lat}_{ij}$\;
}
\If{do inhibition?}{
    \For{$y_{ij} \in Y$}{
        $inhibition \longleftarrow \sum_{kl \in \mathcal{N}(ij,r)}
            {\mathbf{1}(y_{kl} \geq y_{ij})}$\;
        $y_{ij}^* \longleftarrow \mathbf{1}(inhibition < sparsity \times \|\mathcal{N}(ij, r)\|)$\Comment*[r]{fire if winner}
    }
    $Y \longleftarrow Y^*$\;
}
\caption{combinePredKernel($fbPreds$, $latPreds$, $blend$)}
\end{procedure}

\paragraph{Prediction Errors} are used in the hierarchy to drive local learning, both in the encoder and decoder.

\begin{procedure}[H]
\DontPrintSemicolon
\KwData{input image $X$}
\KwResult{prediction errors $E_Y$, hidden errors $E_Z$}
\For{each layer $l$}{
    \lIf{bottom layer}{
        $targets \longleftarrow X$
    }
    \lElse{
        $targets \longleftarrow D$
    }
    $E_Y \longleftarrow targets - \bar{Y}$\Comment*[r]{elementwise}
    $E_Z \longleftarrow propagateErrors(E_Y, \bar{E_Z}, \mathbf{W}_{lat}, r_{lat})$\;
}
\caption{hierarchyErrors($inputs$)}
\end{procedure}

\paragraph{propagateErrors} reverses the decoding transformation to derive the errors in the encoder hidden states.

\begin{procedure}[H]
\DontPrintSemicolon
\KwData{prediction errors $E^Y$, previous hidden errors $\bar{E^Z}$, prediction weights $\mathbf{W}_{lat}$, prediction radius $r_{lat}$}
\KwResult{new hidden errors $E^Z$}
\For{$\epsilon^Z_{ij} \in E^Z$}{
    $(k_0, l_0) \longleftarrow projection_{Z\rightarrow Y}(i,j)$\;
    $\epsilon^Z_{ij} \longleftarrow \bar{\epsilon}^Z_{ij} + \sum_{kl \in         \mathcal{N}(k_0,l_0,r)}{\epsilon^Y_{kl} \mathbf{w}_{klij}}$\;
}
\caption{propagateErrors($predictionErrors, hiddenErrors$)}
\end{procedure}

\paragraph{Learning} in the hierarchy involves using the prediction errors and the decoder inputs (hidden states or feedback) to update decoder weights, and a Spike-time Dependent Plasticity (STDP)-like rule to update the encoding weights.

\begin{procedure}[H]
\DontPrintSemicolon
\For{each layer $l$}{
    \lIf{top layer}{
        $\bar{Y}_{fb} \longleftarrow \bar{Z}$
    }
    \lElse{
        $\bar{Y}_{fb} \longleftarrow \bar{Y}_{l+1}$
    }
    \If{layer predictive?}{
        $\mathbf{W}_{fb} \longleftarrow learnDecoder(E_Y, \bar{Y}_{fb}, \bar{\mathbf{W}}_{fb}, \alpha_{fb})$\;
        $\mathbf{W}_{lat} \longleftarrow learnDecoder(E_Y, \bar{Z}, \bar{\mathbf{W}}_{lat}, \alpha_{lat})$\;
    }
    \If{layer for input}{
        $\mathbf{W}_{enc} \longleftarrow learnEncoder(S, E_Z, \bar{Z}, \bar{D}, \bar{\mathbf{W}}_{enc}, \alpha_{enc})$\;
    }
    $B \longleftarrow learnBiases(S, \bar{B}, \alpha_{bias})$\;
}
\caption{hierarchyLearn()}
\end{procedure}

\paragraph{learnDecoder()} updates the decoder weights by combining the prediction error on the output with the input to that weighted connection.

\begin{procedure}[H]
\DontPrintSemicolon
\KwData{prediction error $E^Y$, inputs $\bar{Z}$, weights $\bar{\mathbf{W}}$, decay $\alpha$}
\KwResult{new prediction weights $\mathbf{W}$}
\For{$w_{ijkl} \in \mathbf{W}$}{
    $w_{ijkl} \longleftarrow \bar{w}_{ijkl} + \alpha \epsilon^Y_{kl} \bar{z}_{ij}$\;
}
\caption{learnDecoder($predErrors, prevInputs, weights, \alpha$)}
\end{procedure}

\paragraph{learnEncoder} uses a STDP-like mechanism to update the encoder weights, combining the hidden errors, the hidden state, and the derived inputs at each end of the weight's connection. The second component of the weights tensor is a trace of past correlations of hidden states and inputs. The factor $scale$ uses the second component of the stimulus tensor, which is the squared norm of the incoming encoder weights.

\begin{procedure}[H]
\DontPrintSemicolon
\KwData{stimulus $S$, hidden errors $E^Z$, hidden states $\bar{Z}$, derived input $\bar{D}$, weights $\bar{\mathbf{W}}_{enc}$, $\alpha_{enc}$}
\KwResult{new encoder weights $\mathbf{W}$}
\For{$w_{ijkl} \in \mathbf{W}$}{
    $scale \longleftarrow \frac{1}{\sqrt{max(10^{-4}, s^2_{kl})}}$\;
    $w^1_{ijkl} \longleftarrow \bar{w}^1_{ijkl} * scale + \alpha  \epsilon^Z_{kl} \bar{z}_{kl} (\bar{d}^1_{ij} - \bar{d}^2_{ij})$\;
    $w^2_{ijkl} \longleftarrow \bar{w}^2_{ijkl} * \gamma + (1 - \gamma) \bar{z}_{kl}(\bar{d}^1_{ij} - \bar{d}^2_{ij})$\Comment*[r]{trace}
}
\caption{learnEncoder($stimuli, errors, states, derived, W, \alpha$)}
\end{procedure}

\paragraph{} The biases may be updated in several manners, in a way such that the states of a unit have a desired lifetime sparsity. One such method is to updated the biases such that the average stimulus received by a unit approaches zero

\begin{procedure}
\DontPrintSemicolon
\KwData{stimulus $S$, previous biases $\bar{B}$, $\alpha_{bias}$}
\KwResult{updated biases $B$}
\For{$b_{ij} \in B$}{
    $b_{ij} \longleftarrow (1 - \alpha) \bar{b}_{ij} - \alpha s_{ij}$\;
}
\caption{learnBiases($S, \bar{B}, \alpha_{bias}$)}
\end{procedure}

\subsection*{S2 Routed Predictive Hierarchy Algorithms}
\label{S2_Algorithms}

The following describes another class of Feynman Machine, the Routed Predictive Hierarchy. This variation differs from the Sparse Predictive Hierarchy in that encoders and decoders interact through modulation as opposed to error propagation. The encoders in what follows are trained separately from the decoding phase, and the decoding phase consists of a decoder network whose states are modulated by the encoders. This causes the encoders to choose sub-networks for the decoder network, simplifying training and more importantly adding spatiotemporal context to predictions.

\paragraph{simStep(inputs, learn)} runs the hierarchy encoding and decoding passes, generates predictions and performs learning optionally.

\begin{procedure}[H]
\DontPrintSemicolon
\KwData{image $inputs$, learning control $learn$}
\Begin{
    $hierarchyEncode(inputs)$\;
    $hierarchyPredict(encoderStates)$\;
    \lIf{learn?}{
        $hierarchyLearn(inputs)$
    }
}
\caption{simStep($inputs$, $learn?$)}
\end{procedure}

\paragraph{encode()} pass input into the hierarchy, and runs the up-pass for each layer.

\begin{procedure}[H]
\DontPrintSemicolon
\KwData{image $inputs$}
\Begin{
    \lIf{whiten}{
       $inputs \longleftarrow whiten(inputs)$
    }
    $visibles \longleftarrow inputs$\;
    \ForEach{encoderLayer $l$}{
        $l.states \longleftarrow l.activate(visibles, \lambda)$\;
        $visibles \longleftarrow l.states$\;
    }
}
\caption{hierarchyEncode($inputs$)}
\end{procedure}

\paragraph{activate()} and its kernels pass the hierarchy's inputs up from layer to layer. 

\begin{procedure}[H]
\DontPrintSemicolon
\KwData{visible input $X$, previous $\bar{D}$, biases $B$, weights $\mathbf{W}_{enc}$, decay $\lambda$}
\KwResult{derived input $D$, stimulus $S$, hidden states $Z$}
\Begin{
    $D \longleftarrow deriveInputKernel(X, \bar{D}, \lambda)$\Comment*[r]{derived input}
    $S \longleftarrow encodeKernel(D, \bar{S}, \mathbf{W}_{enc}, r)$\Comment*[r]{compute stimulus}
    $Z \longleftarrow solveHiddenKernel(S + B, r_{inhibit})$\Comment*[r]{compute hidden states}
}
\caption{activate($visibles$, $\lambda$)}
\end{procedure}

\paragraph{deriveInputKernel} produces a pair of inputs to the encoder, one a copy of the input and the other a moving average of recent inputs.

\begin{procedure}[H]
\DontPrintSemicolon
\KwData{visible inputs $X$, previous output $\bar{D}$, decay $\lambda$}
\KwResult{derived output $D$, $d_{ij}^1$ is the input copied, $d_{ij}^2$ is decayed}
\For{$x_{ij} \in X$}{
    $d_{ij}^1 \longleftarrow x_{ij}$\;
    $d_{ij}^2 \longleftarrow \lambda\bar{d}_{ij}^2 + (1-\lambda)x_{ij}$\Comment*[r]{decay previous}
}
\caption{deriveInputKernel($inputs$, $outputsPrev$, $\lambda$)}
\end{procedure}

\paragraph{} encodeKernel produces the encoder's stimulus, which is a weighted sum of each unit's derived inputs.

\begin{procedure}[H]
\DontPrintSemicolon
\KwData{derived inputs $D$, previous stimulus $\bar{S}$, weights $\mathbf{W}_{enc}$, window radius $r$}
\KwResult{new stimulus $S$}
\For{$\bar{s}_{ij} \in \bar{S}$}{
        $(k_0, l_0) \longleftarrow projection_{S\rightarrow D}(i,j)$\;
        $s_{ij}^1 \longleftarrow \bar{s}_{ij}^1 + 
          \sum_{kl \in \mathcal{N}(k_0, l_0,r)}{
            \mathbf{w}_{ijkl} (d_{kl}^1 - d_{kl}^2)}$\;
    }
\caption{encodeKernel($derivedInput$, $prevStimulus$, $weights$, $radius$)}
\end{procedure}

\paragraph{} solveHiddenKernel implements local inhibition on the activations to produce a sparse binary hidden state.

\begin{procedure}[H]
\DontPrintSemicolon
\KwData{activations $A$, radius $r$}
\KwResult{new hidden states $Z$}
\For{${a}_{ij} \in A$}{
    $inhibition \longleftarrow \sum_{kl \in \mathcal{N}(ij,r)}
        {\mathbf{1}(a_{kl} \geq a_{ij})}$\;
    $z_{ij} \longleftarrow \mathbf{1}(inhibition \leq sparsity \times \|\mathcal{N}(ij, r)\|)$\Comment*[r]{fire if winner}    
}
\caption{solveHiddenKernel($activations$, $radius$)}
\end{procedure}

\paragraph{Prediction} is handled by a standard feed-forward deep neural network, with the caveat that the activations are modulated by the states of the encoder portion of the system. This prediction network is trained via backpropagation or other means while taking into account the modulating states $Z$ from the encoder hierarchy. This modulation forms subnetworks of the predictor network that are specific to a spatiotemporal state, allowing for improved generalization capabilities as well as fast online training. As the modulation introduces a nonlinearity into the system, the predictor network can be a linear function approximator, simplifying the credit assignment problem due to lack of a vanishing gradient.

Decoding uses a standard feedforward deep neural network operated upside down (from upper layers towards the input layer). Information passes down the hierarchy, as each layer generates predictions by decoding its input from higher layer decoder predictions, modulated by its encoder's hidden states (ie $z^{mod}_{ij} \longleftarrow z^{fb}_{ij}z_{ij}$).

\begin{procedure}[H]
\DontPrintSemicolon
\KwResult{prediction $Y_l$ for each layer}
 \For{each layer $l$ in reverse order}{
        \lIf{not top layer}{
            $Z_{fb} \longleftarrow Y_{l+1}$
        }
        \lElse{
            $Z_{fb} \longleftarrow Z$
        }
        \If{$l$ predicts}{
            $Z_{mod} \longleftarrow Z_{fb} \odot Z$\Comment*[r]{modulate (multiply elementwise)}
            $Y_l \longleftarrow decodeKernel(Z_{mod}, B_{fb}, \mathbf{W}_{fb}, r_{fb})$\;
        }
}
\caption{decodeHierarchy()}
\end{procedure}

\paragraph{} decodeKernel transforms the modulated feedback inputs to the decoder using a weight matrix, adding biases, and forming a partial prediction via an optional nonlinearity (defaults to the identity function).

\begin{procedure}[H]
\DontPrintSemicolon
\KwData{source $Z$, biases $B$, weights $\mathbf{W}$, window radius $r$}
\KwResult{predictions $Y$}
\For{$y_{ij} \in Y$}{
        $(k_0, l_0) \longleftarrow projection_{Y\rightarrow Z}(i,j)$\;
        $y_{ij} \longleftarrow nonlinearity(b_{ij} + \sum_{kl \in \mathcal{N}(k_0,l_0,r)}{
            z_{kl} \mathbf{w}_{ijkl} })$\;
    }
\caption{decodeKernel($source$, $biases$, $weights$, $radius$)}
\end{procedure}

\paragraph{} hierarchyLearn runs the learning for each layer's encoder.

\begin{procedure}[H]
\DontPrintSemicolon
\For{each layer $l$}{
    \lIf{top layer}{
        $\bar{Y}_{fb} \longleftarrow \bar{Z}$
    }
    \lElse{
        $\bar{Y}_{fb} \longleftarrow \bar{Y}_{l+1}$
    }
    \If{layer for input}{
        $\mathbf{W}_{enc} \longleftarrow learnEncoder(S, E_Z, \bar{Z}, \bar{D}, \bar{\mathbf{W}}_{enc}, \alpha_{enc})$\;
    }
    $B \longleftarrow learnBiases(S, \bar{B}, \alpha_{bias})$\;
}
\caption{hierarchyLearn()}
\end{procedure}

\paragraph{}{Learning} the encoders is performed using a form of STDP.

\begin{procedure}[H]
\DontPrintSemicolon
\KwData{stimulus $S$, hidden states $\bar{Z}$, derived input $\bar{D}$, weights $\bar{\mathbf{W}}_{enc}$, $\alpha_{enc}$}
\KwResult{new encoder weights $\mathbf{W}$}
\For{$w_{ijkl} \in \mathbf{W}$}{
    $w_{ijkl} \longleftarrow \bar{w}_{ijkl} + \alpha  \epsilon^Z_{kl} \bar{z}_{kl} [(\bar{d}^1_{ij} - \bar{d}^2_{ij}) - w_{ijkl}]$\;
}
\caption{learnEncoder($stimuli, errors, states, derived, W, \alpha$)}
\end{procedure}

\paragraph{} Increment the bias when the state is too low (average below the target sparsity), and proportionally decrement it when the state is too high (average above the target sparsity).

\begin{procedure}
\DontPrintSemicolon
\KwData{state $Z$, previous biases $\bar{B}$, $\alpha_{bias}$}
\KwResult{updated biases $B$}
\For{$b_{ij} \in B$}{
    $b_{ij} \longleftarrow \bar{b}_{ij} + \alpha (sparsity - z_{ij})$\;
}
\caption{learnBiases($S, \bar{B}, \alpha_{bias}$)}
\end{procedure}

\section*{Acknowledgments}
The authors gratefully acknowledge the support of Ogma Intelligent Systems Corp which funded this work.


%
%
%


\bibliography{plos}

\begin{thebibliography}{}

\bibitem[\protect\citeauthoryear{Brockman \bgroup et al\mbox.\egroup
  }{2016}]{openai}
Brockman, G.; Cheung, V.; Pettersson, L.; Schneider, J.; Schulman, J.; Tang,
  J.; and Zaremba, W.
\newblock 2016.
\newblock {OpenAI Gym}.
\newblock {\em CoRR} abs/1606.01540.

\bibitem[\protect\citeauthoryear{Byrne}{2015a}]{byrne2015paCLA}
Byrne, F.
\newblock 2015a.
\newblock {E}ncoding {R}eality: {P}rediction-{A}ssisted {C}ortical {L}earning
  {A}lgorithm in {H}ierarchical {T}emporal {M}emory.
\newblock {\em arXiv preprint arXiv:1509.08255}.

\bibitem[\protect\citeauthoryear{Byrne}{2015b}]{byrne2015symphony}
Byrne, F.
\newblock 2015b.
\newblock Symphony from synapses: Neocortex as a universal dynamical systems
  modeller using hierarchical temporal memory.
\newblock {\em arXiv preprint arXiv:1512.05245}.

\bibitem[\protect\citeauthoryear{Cadieu \bgroup et al\mbox.\egroup
  }{2014}]{cadieu2014deep}
Cadieu, C.~F.; Hong, H.; Yamins, D.~L.; Pinto, N.; Ardila, D.; Solomon, E.~A.;
  Majaj, N.~J.; and DiCarlo, J.~J.
\newblock 2014.
\newblock {Deep neural networks rival the representation of primate IT cortex
  for core visual object recognition}.
\newblock {\em PLoS Comput Biol} 10(12):e1003963.

\bibitem[\protect\citeauthoryear{Carmantini \bgroup et al\mbox.\egroup
  }{2016}]{carmantini2016modular}
Carmantini, G.~S.; Desroches, M.; Rodrigues, S.; et~al.
\newblock 2016.
\newblock A modular architecture for transparent computation in recurrent
  neural networks.
\newblock {\em arXiv preprint arXiv:1609.01926}.

\bibitem[\protect\citeauthoryear{Decorsi{\`e}re \bgroup et al\mbox.\egroup
  }{2015}]{decorsiere2015inversion}
Decorsi{\`e}re, R.; S{\o}ndergaard, P.~L.; MacDonald, E.~N.; and Dau, T.
\newblock 2015.
\newblock Inversion of auditory spectrograms, traditional spectrograms, and
  other envelope representations.
\newblock {\em IEEE/ACM Transactions on Audio, Speech, and Language Processing}
  23(1):46--56.

\bibitem[\protect\citeauthoryear{Goodfellow, Bengio, and
  Courville}{2016}]{Goodfellow-et-al-2016-Book}
Goodfellow, I.; Bengio, Y.; and Courville, A.
\newblock 2016.
\newblock Deep learning.
\newblock Book in preparation for MIT Press.

\bibitem[\protect\citeauthoryear{Hawkins and Ahmad}{2015}]{hawkins2015neurons}
Hawkins, J., and Ahmad, S.
\newblock 2015.
\newblock Why neurons have thousands of synapses, a theory of sequence memory
  in neocortex.
\newblock {\em arXiv preprint arXiv:1511.00083}.

\bibitem[\protect\citeauthoryear{Hillis}{1999}]{Hillis:1999:RFC:304763.305699}
Hillis, W.~D.
\newblock 1999.
\newblock Feynman and computation.
\newblock Cambridge, MA, USA: Perseus Books.
\newblock chapter Richard Feynman and the Connection Machine,  257--265.

\bibitem[\protect\citeauthoryear{Lavin and Ahmad}{2015}]{lavin2015evaluating}
Lavin, A., and Ahmad, S.
\newblock 2015.
\newblock {Evaluating Real-Time Anomaly Detection Algorithms--The Numenta
  Anomaly Benchmark}.
\newblock In {\em 2015 IEEE 14th International Conference on Machine Learning
  and Applications (ICMLA)},  38--44.
\newblock IEEE.

\bibitem[\protect\citeauthoryear{LeCun, Bengio, and
  Hinton}{2015}]{lecun2015deep}
LeCun, Y.; Bengio, Y.; and Hinton, G.
\newblock 2015.
\newblock Deep learning.
\newblock {\em Nature} 521(7553):436--444.

\bibitem[\protect\citeauthoryear{Lorenz}{1963}]{lorenz1963}
Lorenz, E.~N.
\newblock 1963.
\newblock Deterministic nonperiodic flow.
\newblock {\em Journal of the Atmospheric Sciences} 20(2):130--141.

\bibitem[\protect\citeauthoryear{Makhzani and Frey}{2013}]{makhzani2013k}
Makhzani, A., and Frey, B.~J.
\newblock 2013.
\newblock k-sparse autoencoders.
\newblock {\em CoRR} abs/1312.5663.

\bibitem[\protect\citeauthoryear{McCulloch and
  Pitts}{1943}]{mcculloch1943logical}
McCulloch, W.~S., and Pitts, W.
\newblock 1943.
\newblock A logical calculus of the ideas immanent in nervous activity.
\newblock {\em The bulletin of mathematical biophysics} 5(4):115--133.

\bibitem[\protect\citeauthoryear{Neumann}{1958}]{neumann1958computer}
Neumann, J.~v.
\newblock 1958.
\newblock The computer and the brain.

\bibitem[\protect\citeauthoryear{Poincar{\'e}}{1881}]{poincare1881}
Poincar{\'e}, H.
\newblock 1881.
\newblock {M{\'e}moire sur les courbes d{\'e}finies par une {\'e}quation
  diff{\'e}rentielle (I)}.
\newblock {\em Journal de math{\'e}matiques pures et appliqu{\'e}es}
  7:375--422.

\bibitem[\protect\citeauthoryear{Poincar{\'e}}{1992}]{poincare1992new}
Poincar{\'e}, H.
\newblock 1992.
\newblock {\em New methods of celestial mechanics}, volume~13.
\newblock Springer Science \& Business Media.

\bibitem[\protect\citeauthoryear{Rasmus \bgroup et al\mbox.\egroup
  }{2015}]{rasmus2015semi}
Rasmus, A.; Berglund, M.; Honkala, M.; Valpola, H.; and Raiko, T.
\newblock 2015.
\newblock Semi-supervised learning with ladder networks.
\newblock In {\em Advances in Neural Information Processing Systems},
  3546--3554.

\bibitem[\protect\citeauthoryear{Sauer}{2006}]{Sauer:2006}
Sauer, T.~D.
\newblock 2006.
\newblock {A}ttractor reconstruction.
\newblock Revision 91017.

\bibitem[\protect\citeauthoryear{Siedenburg and
  D{\"o}rfler}{2012}]{siedenburg2012audio}
Siedenburg, K., and D{\"o}rfler, M.
\newblock 2012.
\newblock Audio denoising by generalized time-frequency thresholding.
\newblock In {\em Audio Engineering Society Conference: 45th International
  Conference: Applications of Time-Frequency Processing in Audio}.
\newblock Audio Engineering Society.

\bibitem[\protect\citeauthoryear{Stowell and
  Turner}{2015}]{stowell2015denoising}
Stowell, D., and Turner, R.~E.
\newblock 2015.
\newblock Denoising without access to clean data using a partitioned
  autoencoder.
\newblock {\em CoRR} abs/1509.05982.

\bibitem[\protect\citeauthoryear{Strogatz}{2014}]{strogatz2014nonlinear}
Strogatz, S.~H.
\newblock 2014.
\newblock {\em {N}onlinear {d}ynamics and {c}haos: with applications to
  physics, biology, chemistry, and engineering}.
\newblock Westview press.

\bibitem[\protect\citeauthoryear{Takens}{1981}]{takens1981detecting}
Takens, F.
\newblock 1981.
\newblock {\em Detecting strange attractors in turbulence}.
\newblock Springer.

\bibitem[\protect\citeauthoryear{Thiede \bgroup et al\mbox.\egroup
  }{2000}]{thiede2000peaq}
Thiede, T.; Treurniet, W.~C.; Bitto, R.; Schmidmer, C.; Sporer, T.; Beerends,
  J.~G.; and Colomes, C.
\newblock 2000.
\newblock {PEAQ - The ITU standard for objective measurement of perceived audio
  quality}.
\newblock {\em Journal of the Audio Engineering Society} 48(1/2):3--29.

\bibitem[\protect\citeauthoryear{Whitney}{1936}]{whitney1936differentiable}
Whitney, H.
\newblock 1936.
\newblock Differentiable manifolds.
\newblock {\em Annals of Mathematics}  645--680.

\bibitem[\protect\citeauthoryear{Yu, Mallat, and Bacry}{2008}]{yu2008audio}
Yu, G.; Mallat, S.; and Bacry, E.
\newblock 2008.
\newblock Audio denoising by time-frequency block thresholding.
\newblock {\em IEEE Transactions on Signal processing} 56(5):1830--1839.

\bibitem[\protect\citeauthoryear{Zhu, Beauregard, and Wyse}{2007}]{zhu2007real}
Zhu, X.; Beauregard, G.~T.; and Wyse, L.~L.
\newblock 2007.
\newblock {Real-time signal estimation from modified short-time Fourier
  transform magnitude spectra}.
\newblock {\em IEEE Transactions on Audio, Speech, and Language Processing}
  15(5):1645--1653.

\end{thebibliography}

\end{document}